\documentclass[runningheads]{llncs}

\usepackage{eccv}

\usepackage{booktabs}
\usepackage[table]{xcolor}
\definecolor{best}{RGB}{0,150,0}
\definecolor{second}{RGB}{230,120,0}
\usepackage{pifont}
\newcommand{\cmark}{\textcolor{green}{\ding{51}}}
\newcommand{\xmark}{\textcolor{red}{\ding{55}}}

\usepackage{eccvabbrv}
\usepackage{xcolor}
\usepackage{graphicx}
\usepackage{booktabs}

\usepackage[accsupp]{axessibility}  

\usepackage{hyperref}
\hypersetup{
  pdftitle={Geometric Foundation Model Distillation for Efficient Lunar 3D Reconstruction},
  pdfauthor={Clementine Grethen, Florient Chouteau, Geraldine Morin, Simone Gasparini}
}

\definecolor{best}{rgb}{0.0, 0.5, 0.0}
\definecolor{second}{rgb}{0.7, 0.25, 0.0}
\newcommand{\best}[1]{\textcolor{best}{\textbf{#1}}}
\newcommand{\second}[1]{\textcolor{second}{#1}}

\usepackage{orcidlink}

\usepackage[per-mode=symbol]{siunitx}
\usepackage{xspace}
\usepackage{xcolor}
\usepackage[normalem]{ulem}
\usepackage[capitalize,noabbrev]{cleveref}

\makeatletter
\DeclareRobustCommand\onedot{\futurelet\@let@token\@onedot}
\def\@onedot{\ifx\@let@token.\else.\null\fi\xspace}
\DeclareRobustCommand\nodot{\futurelet\@let@token\@nodot}
\def\@nodot{\ifx\@let@token.\else~\null\fi\xspace}

\newfont{\eaddfnt}{phvr8t at 12pt}

\def\eg{\emph{e.g}\onedot} 
\def\vs{vs\onedot}

\def\cf{\emph{cf}\onedot} 
 \def\vs{\emph{vs}\onedot}

\DeclareSIUnit\px{px}
\newcommand{\myparagraph}[1]{\noindent\textbf{#1}}

\begin{document}

\title{Geometric Foundation Model Distillation \\ for Efficient Lunar 3D Reconstruction}

\titlerunning{Geometric Foundation Model Distillation}

\author{
Clémentine Grethen\inst{1}\orcidID{0009-0009-3695-1717} \and
Florient Chouteau\inst{2} \and
Géraldine Morin\inst{1}\orcidID{0000-0003-0925-3277} \and
Simone Gasparini\inst{1}\orcidID{0000-0001-8239-8005}
}

\authorrunning{C.~Grethen}

\institute{
\inst{1} IRIT, University of Toulouse, France \and
\inst{2} Airbus Defence and Space, Toulouse, France }
\maketitle

\begin{abstract}

\begin{figure}
    \centering
    \includegraphics[width=1\linewidth]{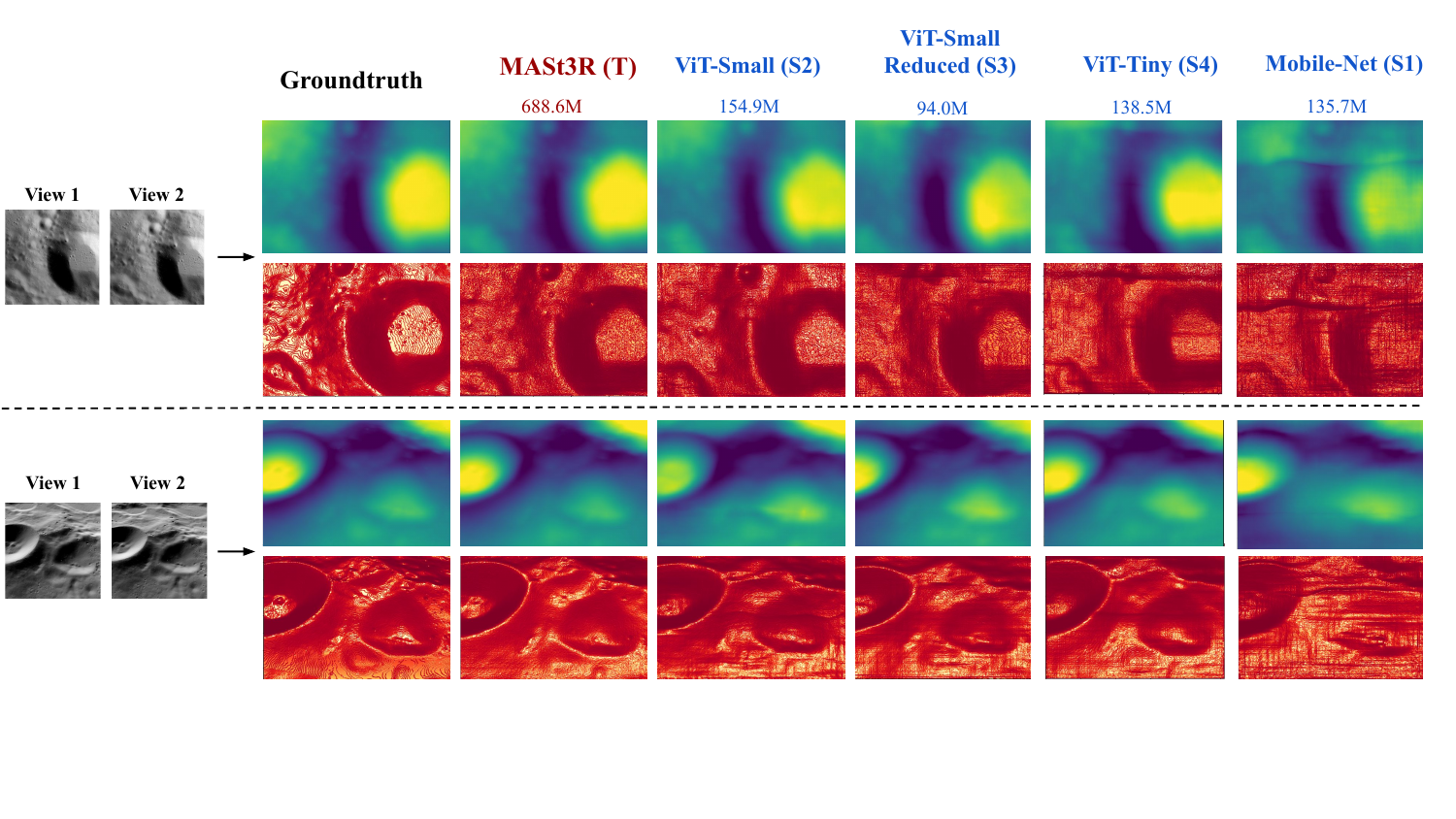}
    \caption{
     Our \textbf{\textcolor[HTML]{1155CC}{student}} models, distilled from the 
\textbf{\textcolor[HTML]{8B0000}{teacher}} (MASt3R, $688.6$M parameters), 
recover depth maps (top) and surface slopes (bottom) for view~1 of two 
uncalibrated lunar stereo pairs. Students are ordered left to right from best to weakest: S2 
(ViT-Small, $154.9$M, $4.4\times$ smaller) closely matches the 
\textbf{\textcolor[HTML]{8B0000}{teacher}} in both depth and slope; S3 
(reduced decoder, $94.9$M, $7.3\times$ smaller) remains competitive; S4 
(ViT-Tiny, $138.5$M) shows slight degradation; while S1 (MobileNet, $135.7$M) 
exhibits clear slope artifacts.}
    \label{fig:teasing}
\end{figure}

Large 3D foundation models such as MASt3R achieve state-of-the-art stereo reconstruction but are computationally demanding for deployment under strict hardware constraints---a critical limitation in domains such as planetary exploration, where onboard computing is severely restricted.
We study how far such models can be compressed through knowledge distillation, using lunar stereo reconstruction as a challenging and practically relevant case study.
Starting from a $688$M-parameter MASt3R teacher fine-tuned on lunar imagery, we distill its dense geometric predictions into a family of lightweight students spanning different encoder types (CNN \vs ViT), decoder widths and depths, and training strategies.
To bridge the dimensional mismatch between teacher and student, we propose a structured SVD-based initialization that projects the teacher's decoder weights into the student's smaller latent space, yielding a warm start that significantly improves convergence and final performance. 
Based on our results on lunar data, we can obtain a distilled student that retains most of teacher's reconstruction accuracy while reducing the model size up to $7\times$, and even outperforms a baseline trained directly with sparse ground-truth annotations.
Beyond compression, our study highlights both principles and practical insights for distilling geometric foundation models: a convolutional encoder underperforms transformer-based alternatives (though pretraining availability remains a confounding factor), preserving encoder capacity is more critical than maintaining a large decoder, feature-level distillation consistently outperforms output-only supervision, and SVD-based initialization improves optimisation stability.
These findings provide practical guidelines for deploying 3D reconstruction models in resource-constrained environments. Code is publicly available at \url{https://clementinegrethen.github.io/publications/ECCV.html}.

\end{abstract}

\section{Introduction}

Recent advances in large-scale learning-based approaches have significantly advanced multi-view 3D reconstruction. In particular, recent 3D foundation models—models capable of inferring globally consistent geometry directly from images—have demonstrated state-of-the-art performance. Architectures such as DUSt3R~\cite{wang2023dust3r}, MASt3R~\cite{leroy2024grounding}, and VGGT~\cite{wang2025vggt} show that dense 3D geometry can be recovered from image pairs without explicit camera calibration. By leveraging massive pre-training and high-capacity transformer backbones (ViTs~\cite{dosovitskiy2021imageworth16x16words}), these models achieve state-of-the-art performance across a wide range of terrestrial datasets. However, this performance comes at a considerable cost: such models typically contain hundreds of millions to over a billion parameters and require substantial memory and computational resources for both training and inference.
While such capacity is effective for general-purpose 3D reasoning, it raises an important question: \emph{is this level of complexity necessary for targeted, mission-specific applications?}

More broadly, the increasing reliance on extremely large models has created a growing gap between cutting-edge research systems and practical deployment environments.
This disparity between high-capacity research models and real-world applications highlights a divergence between innovation and accessibility.
While recent advances in multi-view geometry rely on massive computational resources, the size and inference cost of such models make them difficult to deploy in practice.
In particular, many real-world platforms lack the memory and computational capacity required to run networks of this scale, limiting their applicability in scenarios where 3D reconstruction is most valuable, such as robotics or autonomous navigation~\cite{wu2022tinyvit}.

For example, these limitations become particularly evident in planetary exploration. In lunar missions, onboard processing units on landers or rovers operate under strict memory, power-consumption, and latency constraints \cite{Getchius2024}.
At the same time, accurate 3D reconstruction is essential for critical tasks such as safe landing, hazard detection, descent navigation, and terrain analysis~\cite{grethen2025adaptingstereovisionobjects}.
The lunar environment further complicates the problem due to specific conditions rarely encountered in standard training datasets, including low-texture surfaces, extreme illumination conditions, and limited parallax during descent trajectories \cite{Getchius2024,kumar2024moonmetasynclunarimageregistration}. 
Although dedicated datasets such as StereoLunar~\cite{grethen2025adaptingstereovisionobjects} have demonstrated the feasibility of learning-based approaches, the size and inference cost of modern transformer-based models make their deployment in such settings particularly challenging.

These observations motivate an alternative approach.
Instead of designing increasingly large universal models, an alternative approach is to compress and specialize existing foundation models for specific operational contexts.
Knowledge distillation~\cite{Gou2021} provides a systematic approach for this objective: a lightweight student network is trained to reproduce the geometric predictions of a large teacher model, inheriting its representational priors while significantly reducing computational cost.

Despite its success in language modeling~\cite{sanh2020distilbertdistilledversionbert,xu2024surveyknowledgedistillationlarge}, distillation for dense geometric foundation models remains less explored.
Several important questions, therefore, arise:

\begin{itemize}
    \item How should a student's architecture be designed for multi-view geometry?
    \item Is a convolutional encoder sufficient, or are vision transformers still required?
    \item How to trade off between model size and reconstruction accuracy?
\item What training strategies are most effective when distilling large transformer-based teachers into lightweight student networks?
\end{itemize}

In this work, we study how far a large geometric foundation model can be compressed for a targeted operational domain, using lunar descent and orbital imagery as a challenging and practically relevant case study.
Starting from a MASt3R-based teacher fine-tuned on lunar data ($688$M parameters), we distill its dense geometric predictions into a family of lightweight students employing different encoder types (CNN \vs ViT), decoder width/depth, and training choices (\eg, encoder freezing, feature-level supervision). We select MASt3R as the teacher due to its favorable trade-off between geometric accuracy and computational efficiency among other 3D foundation models.
While our experiments focus on the Moon, the proposed recipe, distilling a domain-adapted 3D teacher into efficient students, applies broadly to other deployment-constrained domains.

Experimentally, our ViT-based students compress the $688.6$M-parameter teacher by $4.4\times$ to $7.3\times$ while delivering a consistent $\sim\!2\times$ inference speedup.
Despite this compression, the best student (S2, ViT-Small) preserves reconstruction quality with only $15\%$ higher Chamfer error than the teacher (see \cref{fig:teasing}).
Its reduced-decoder variant (same encoder, decoder reduced from 512/6 to 256/4) is $7.3\times$ smaller, with only marginal additional degradation, suggesting that much of the decoder capacity is redundant for this task (see \cref{fig:teasing}).
Moreover, distillation with teacher pseudo-ground truth outperforms standard ground-truth supervision, reducing Chamfer error by $\approx\!18\%$, confirming that the teacher provides a richer dense supervisory signal than available labels alone.

{Our main contributions are:}
\begin{itemize}
    \item  A general knowledge distillation framework for stereo 3D reconstruction that transfers a $688$M-parameter teacher model into compact student networks without requiring pose annotations, using teacher predictions as dense pseudo-ground truth. We demonstrate its effectiveness on lunar stereo reconstruction.
    \item A quantified study of the impact of architecture choices on geometry and pose estimation: (a) encoder choice, (b) encoder fine-tuning strategy (frozen \vs trainable), and (c) targeted decoder compression.
    \item A structured SVD-based initialization for compressing transformer decoder weights, improving convergence and final performance.
    \item A training-strategy ablation isolating key distillation ingredients (output supervision, feature alignment, and SVD warm-start) and measuring their effect on reconstruction and pose metrics.

\end{itemize}

The remainder of the paper is organized as follows: \cref{sec:related} reviews related work, \cref{sec:method} presents our distillation framework and initialization strategy, and \cref{sec:experiments} reports experimental results and ablation studies.

\section{Related work}
\label{sec:related}
\myparagraph{Foundation models for dense Multi-View 3D Reconstruction.}
Classical Structure-from-Motion (SfM)\cite{1979,Hartley2004} pipelines reconstruct sparse 3D geometry by matching features across multiple views and estimating camera motion, but remain limited in their ability to produce dense, globally consistent reconstructions.
Recent end-to-end models such as DUSt3R~\cite{wang2023dust3r} and MASt3R~\cite{leroy2024grounding} directly regress geometry and camera parameters from stereo image pairs, achieving strong performance on terrestrial scenes.
However, when extended to multi-view settings, they typically require an additional global alignment stage to ensure cross-view consistency.

VGGT~\cite{wang2025vggt} addresses this limitation by enabling fully end-to-end dense reconstruction from image sequences. It jointly predicts depth, point clouds, and camera parameters across multiple views, producing geometrically consistent representations without explicit global alignment.
While highly effective, VGGT relies on a very large transformer-based architecture with over $1$ billion parameters.
These recent 3D foundation models range from roughly $650$M to over $1$B parameters and require massive training infrastructure, often exceeding $100$ A100 GPUs.
Such computational demands impose substantial memory, energy, and financial costs, limiting reproducibility and slowing research iteration.
Although these large-scale models demonstrate strong generalization across diverse scenes, their capacity often exceeds the requirements of task-specific applications such as navigation or domain-constrained reconstruction, where efficiency and reliability are more critical than universal coverage.
Moreover, they cannot always generalize to highly specific domains (\eg, space imagery) because the training datasets do not sufficiently capture such specialized conditions~\cite{grethen2025adaptingstereovisionobjects}.
This mismatch creates an infrastructure bottleneck, with model scale surpassing practical deployment needs. 

\myparagraph{Knowledge Distillation.}
Knowledge distillation (KD) \cite{Gou2021} is a widely used paradigm for transferring the capabilities of large neural networks into compact, efficient models. Originally introduced to leverage the ``dark knowledge'' contained in a teacher's soft predictions \cite{hinton2015distilling}, KD enables a student network to learn richer supervisory signals than hard labels alone, capturing inter-class correlations that improve generalization \cite{touvron2021training}.
This paradigm is particularly attractive where memory, latency, and energy consumption are critical constraints.
Distillation can operate at different supervision levels: response-based distillation trains the student to reproduce the teacher's final outputs, while feature-level distillation further aligns intermediate representations, encouraging the student to mimic internal reasoning processes \cite{Gou2021}.

Recent works have explored distillation or adaptation of large 3D vision models, but with objectives fundamentally different from ours. \cite{vuong2025improvingroboticmanipulationefficient} distill VGGT into a lighter encoder for robotic manipulation, without any 3D prediction head. \cite{guo2025gladgeometriclatentdistillation} align a vision-language-action model with VGGT features for cross-modal grounding rather than compression. 
\cite{StreamVGGT} modifies VGGT's attention mechanism for temporal streaming, without size constraints.
In contrast, our approach compresses both encoder and decoder, initializes the compressed decoder via structured SVD truncation of teacher weights, and preserves the full dual-decoder 3D prediction pipeline. None of these works tackle this joint compression-and-initialization problem for stereo 3D reconstruction.
Closer to our setting, \cite{dutt2026multiview} explores knowledge distillation for stereo 3D geometry, using DUSt3R as teacher. 
However, their evaluation is confined to a single indoor dataset (and per-scene optimization) with only three student architectures (CNN and ViT), without an ablation of 
training strategies or weight initialisation, leaving open the question of what drives compression quality.

DUNE~\cite{sariyildiz2025dune} distils three heterogeneous teachers (MASt3R, DINOv2 \cite{oquab2023dinov2}, Multi-HMR \cite{baradel2024multi}) into a single encoder. While it produces geometry-aware features and can serve as a backbone for downstream tasks~\cite{leblanc2026distill3r}, DUNE does not perform end-to-end distillation of a 3D foundation model, nor does it address structured compression: it focuses solely on encoder fusion, with no 3D reconstruction pipeline or systematic variation of student capacity.
In contrast, we compress MASt3R (encoder, dual cross-view decoder, and prediction heads) within a single distillation 
framework. 
Our goal is to identify which components of large geometric transformers are essential for maintaining reconstruction fidelity, and which can be safely compressed for lightweight, specific task deployment.

\myparagraph{Moon 3D Reconstruction.} Lunar 3D reconstruction is critical for navigation, hazard assessment, and terrain analysis in the context of descent and landing, but remains challenging due to low texture, repetitive patterns, and extreme illumination changes~\cite{Getchius2024}.
Classical photometric methods and SfM pipelines often degrade in this setting because correspondences are hard to establish, and descent trajectories may provide limited parallax~\cite{kumar2024moonmetasynclunarimageregistration}.
Recent work has shown that large 3D foundation models can be adapted to lunar imagery, as shown in~\cite{grethen2025adaptingstereovisionobjects}, where MASt3R has been successfully fine-tuned on the StereoLunar dataset.
While such adaptations improve robustness, the resulting transformer-scale architectures remain costly in terms of memory and latency, limiting their deployment under strict onboard computing resources. 
In this work, we leverage this lunar-adapted MASt3R as an ideal teacher model to distill compact student networks, serving as a proof-of-concept for domain-specific distillation. 
This setup provides a controlled setting to study how different distillation strategies affect reconstruction fidelity and demonstrates a framework that could be applied to other specialized datasets and tasks.

\section{Method}
\label{sec:method}

We propose a multi-student knowledge distillation framework that compresses a large MASt3R teacher into a family of lightweight students, each balancing efficiency and accuracy. Instead of predefining variants, we systematically reduce MASt3R components while preserving geometric fidelity.
The MASt3R teacher, adapted to lunar imagery as described in \cite{grethen2025adaptingstereovisionobjects}, generates pseudo-ground truth 3D reconstructions to supervise all students without camera pose annotations. Students retain MASt3R’s decoder architecture but vary in compression strategy, including encoder type (CNN \vs ViT), decoder width and depth, and whether the encoder weights are frozen or updated during training. 
This setup enables systematic evaluation of how each architectural and training choice affects lunar 3D reconstruction performance.
\begin{figure}[t]
    \centering
    \includegraphics[width=1\linewidth]{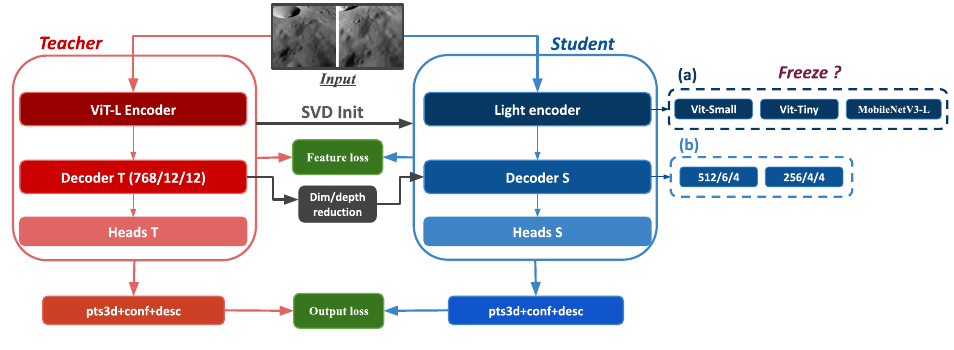}
    \caption{Teacher-student distillation framework. The teacher supervises the student via feature and output losses. (a) Encoders candidates (b) Two compact decoder configurations (dimension/depth/attention heads)}
    \label{fig:student_framework}
\end{figure}

\subsection{Teacher Model}
\label{sec:teacher}

The teacher is a fine-tuned variant of MASt3R~\cite{leroy2024grounding} adapted to lunar orbital imagery.
It has been adapted using the StereoLunar dataset \cite{grethen2025adaptingstereovisionobjects}, which contains stereo pairs of the Moon’s south pole, captured under varying altitudes, baselines, camera motions, and lighting conditions, along with ground-truth 3D scene information. 
As shown in \cref{fig:student_framework}, it follows the CroCo paradigm~\cite{weinzaepfel2022croco}: a shared ViT-Large encoder processes each image of a stereo pair independently, and a cross-attention decoder ($12$ blocks each, dimension $768$, $12$ attention heads) reconstructs dense 3D point maps along with local feature descriptors used for matching. 
The model produces for each pixel, a 3D point expressed in the coordinate frame of the first camera, a confidence score, and a 24-dimensional local descriptor.

Training combines a geometry loss over predicted 3D points and a contrastive matching loss over ground-truth correspondences, jointly encouraging accurate reconstruction and precise feature localisation, as described in \cite{leroy2024grounding}. 
With $688$ million parameters, the teacher serves as a strong but computationally prohibitive baseline, motivating the knowledge distillation framework described in the following section.

\subsection{Student Architectures}
We design the student architectures to mirror the teacher's global reasoning capabilities while fitting within consumer hardware constraints and significantly reducing the parameter count. 
As shown in \cref{fig:student_framework}, we retain the meta-architecture of the teacher but replace heavy components with lightweight alternatives to identify the most suitable configuration for onboard deployment.
    
All student networks share the same modular structure: (a) a lightweight encoder extracts features, a linear projection with group normalisation projects them into the decoder embedding space, and  (b) a smaller cross-view decoder produces the dense geometric predictions.
    
We evaluate three encoder backbones: (a1) a CNN baseline (MobileNetV3-Large \cite{Howard_2019_ICCV}), (a2) a self-supervised ViT (DINOv2 ViT-S/14 \cite{oquab2023dinov2}), and (a3) a smaller ViT (ViT-Tiny/16 \cite{dosovitskiy2021imageworth16x16words}), to cover a spectrum of efficiency and representation quality.
    
On the decoder side, we consider two compact configurations: (b1) a reduced decoder with embedding dimension/depth $(512,6)$, and a more aggressively compressed variant (b2) $(256,4)$, to test how far cross-view reasoning can be shrunk while maintaining reconstruction fidelity.
    
For all student decoders, we fix the number of attention heads to $h=4$, following the multi-head attention formulation introduced by \cite{vaswani2017attention}. 
This results in a per-head dimension of $128$ for $d=512$ and $64$ for $d=256$, preserving sufficient representational capacity per head while keeping attention computation lightweight. 
Finally, the prediction heads (for pointmap, descriptors, and confidence maps) are shared across all student configurations and kept identical to the teacher's (\cf \cref{fig:student_framework}).
    
\subsection{Initialization Strategy}

Initializing student decoder weights randomly leads to slow convergence, since the student must first learn from scratch to produce geometrically meaningful outputs before it can benefit from fine-grained teacher supervision. 
To address this, we propose a structured initialization based on truncated Singular Value Decomposition (SVD). 
While inspired by the low-rank initialization in \cite{sy2024lillama}, we extend this approach to bridge the dimensional mismatch between teacher and student weights, effectively projecting the teacher’s knowledge into the student's smaller latent space.

\myparagraph{Meaningful dimension reduction.} 
Any weight matrix $\mathbf{W}_T \in \mathbb{R}^{d_T^{\text{out}} \times d_T^{\text{in}}}$ of the teacher implements a linear transformation that
can be decomposed via its Singular Value Decomposition (SVD) as:
\begin{equation}
    \mathbf{W}_T = \mathbf{U} \, \boldsymbol{\Sigma} \, \mathbf{V}^\top,
    \quad
    \mathbf{U} \in \mathbb{R}^{d_T^{\text{out}} \times k}, \;
    \boldsymbol{\Sigma} = \operatorname{diag}(\sigma_1, \ldots, \sigma_k), \;
    \mathbf{V}^\top \in \mathbb{R}^{k \times d_T^{\text{in}}},
\end{equation}
where $k = \min(d_T^{\text{out}}, d_T^{\text{in}})$ and the singular values
are ordered $\sigma_1 \geq \sigma_2 \geq \cdots \geq \sigma_k \geq 0$.
The columns of $\mathbf{V}$ define an orthonormal basis of \emph{input directions}, those of $\mathbf{U}$ a corresponding basis of \emph{output directions}, and each $\sigma_i$ quantifies the amplification along the $i$-th direction pair.
A large singular value signals a direction that the weight matrix acts on strongly, while $\sigma_i \approx 0$ indicates a direction essentially ignored by the learned transformation. The singular values thus provide a natural ranking of the importance of each direction in the teacher's weight space.

\myparagraph{Truncated approximation.}
Given a student's weight of shape
$d_S^{\text{out}} \times d_S^{\text{in}}$ with
$d_S^{\text{out}} \leq d_T^{\text{out}}$ and
$d_S^{\text{in}} \leq d_T^{\text{in}}$, we initialize $\mathbf{W}_S$ by retaining the $r = \min(d_S^{\text{out}}, d_S^{\text{in}})$ most important singular components of $\mathbf{W}_T$:
\begin{equation}
    \mathbf{W}_S
    = \mathbf{U}_{[:d_S^{\text{out}},\, :r]} \;
      \operatorname{diag}\!\left(\sigma_1, \ldots, \sigma_r\right) \;
      \mathbf{V}^\top_{[:r,\, :d_S^{\text{in}}]}.
    \label{eq:svd_init}
\end{equation}

%
The SVD orders singular components by decreasing importance, so retaining only the top $r$ discards the directions along which the teacher acts most weakly. 
At the teacher's original dimensions, this rank-$r$ truncation is provably optimal in the Frobenius norm (Eckart--Young--Mirsky theorem)~\cite{eckart1936approximation}. 
\cref{eq:svd_init} further restricts the approximation to the student's smaller dimensions; since the leading singular vectors concentrate the dominant input--output couplings, this extraction preserves the most informative components of the teacher's transformation.
Our SVD-based variant performs the selection in the singular-value basis: we retain the leading singular directions, which capture the most variance in the teacher's weight matrix. This is motivated by the rapid spectral decay observed in Transformer weights \cite{hu2022lora}, which suggests that the dominant singular components encode the most transferable structure.

\myparagraph{Layer mapping.}
Since the student decoder has $S$ blocks while the teacher has $T > S$
blocks, we define a uniform stride mapping from student layer indices to
teacher layer indices:
\begin{equation}
    \phi(i) = \left\lfloor \frac{i \cdot T}{S} \right\rfloor,
    \quad i = 0, \ldots, S-1.
\end{equation}
For our default configuration ($T=12$, $S=6$), this yields
$\phi = [0, 2, 4, 6, 8, 10]$, selecting every other teacher block.
This mapping preserves the hierarchical structure of the decoder: the student's first block inherits low-level cross-view correlations from the teacher's early layers, and its last block inherits high-level geometric representations from the teacher's late layers.

One-dimensional parameters are initialized by direct truncation: $\mathbf{b}_S = \mathbf{b}_T\left[:d_S\right]$.
Unlike weight matrices, these vectors have no directional structure amenable to SVD; we simply retain the first $d_S$ entries.
For convolutional weights $\mathbf{W}_T \in
\mathbb{R}^{C_{\text{out}} \times C_{\text{in}} \times k_h \times k_w}$, we reshape the tensor into a matrix of shape $C_{\text{out}} \times (C_{\text{in}} \cdot k_h \cdot k_w)$, treating each output filter as a single row vector over all its spatial and channel coefficients.

This reshaping preserves the identity of each filter while exposing cross-filter redundancy to the SVD.
We then apply \cref{eq:svd_init} to obtain a compressed matrix of shape $C_{\text{out}}^S \times (C_{\text{in}}^S \cdot k_h \cdot k_w)$, which is reshaped back to $C_{\text{out}}^S \times C_{\text{in}}^S \times k_h \times k_w$. We provide further justifications for our choice of SVD initialization and thoroughly validate the layer mapping strategy through ablation experiments in Sec. 5 of the supplementary material.

\subsection{Implementation Details}
We consider the StereoLunar dataset~\cite{grethen2025adaptingstereovisionobjects}, a physically realistic lunar rendering benchmark inspired by real orbital trajectories, as a benchmark for domain-specific distillation.
The dataset provides diverse lunar imagery with varying baselines, viewpoints, altitudes, and illumination conditions, ensuring a realistic range of geometric and photometric configurations. 
A key design decision is that all student models are trained within the same loop iteration: they share the teacher’s forward pass and the resulting pseudo-ground truth, so a single teacher inference serves multiple students simultaneously, significantly reducing the effective distillation cost per student. 
To improve generalization, student inputs are augmented with color jitter, random crops, grayscale tinting, bilateral filtering, and contrast variations. Models are trained for $50$ epochs on $7$ NVIDIA A100 GPUs using AdamW, with a cosine learning rate decay providing a consistent baseline for performance comparisons while keeping training time manageable.

\subsection{Loss function}
Knowledge distillation methods broadly fall into two categories depending on \emph{where} the teacher's knowledge is extracted from~\cite{Gou2021}.
Output-level (or response-based) distillation trains the
student to reproduce the teacher's final predictions; it is simple to implement, architecture-agnostic, and adds no overhead beyond a single forward pass through the teacher~\cite{hinton2015distilling}.
Feature-level (or hint-based) distillation, introduced
by FitNets~\cite{romero2015fitnets}, additionally aligns intermediate representations between the teacher and student, giving the student access to the teacher's internal reasoning process rather than just its
conclusions.

\myparagraph{Output-level distillation}
We adopt the standard MASt3R geometry loss, which combines a confidence-aware 3D regression term and a cross-view descriptor matching term.
The 3D regression measures the L2 distance between predicted points and pseudo-ground truth points, normalized by the scene’s average depth to ensure scale invariance. 
Confidence predictions are regularized with a logarithmic penalty to prevent trivial solutions.
The descriptor branch is trained with a block-wise InfoNCE \cite{leroy2024grounding} contrastive loss to encourage discriminative cross-view descriptors.
The final geometry loss is then a weighted sum of these two components, $\mathcal{L}_{geo} = \mathcal{L}_{conf} + \beta \, \mathcal{L}_{match}$. 

\myparagraph{Feature alignment loss}
To further guide the student's internal representations, we introduce a
feature-level distillation loss and feature mapping inspired by Depth Anything~\cite{depthanything}. 
After each forward pass, we extract the encoder output features from both the teacher and the student. 
Since the student encoder may have a different channel dimension~$C_s$ than the teacher's~$C_t$, we insert a learnable (trained jointly with the student encoder) linear projection $\mathbf{W} \in \mathbb{R}^{C_t \times C_s}$ that maps student features into the teacher's representation space.
A separate projector is trained for each student's architecture.
Both the projected student features and the (detached) teacher features are $\ell_2$-normalised along the channel axis, yielding unit-norm vectors $\hat{\mathbf{f}}_s$ and $\hat{\mathbf{f}}_t$ at every
spatial location.  
The loss is then a margin-filtered cosine similarity:
\begin{equation}
  \mathcal{L}_{\text{feat}}
  = 1 \;-\;
    \frac{1}{|\mathcal{M}|}
    \sum_{i \in \mathcal{M}}
      \hat{\mathbf{f}}_s^{(i)} \cdot \hat{\mathbf{f}}_t^{(i)},
  \qquad
  \mathcal{M} = \bigl\{\, i \;\big|\;
      \hat{\mathbf{f}}_s^{(i)} \cdot \hat{\mathbf{f}}_t^{(i)} < \alpha
    \bigr\},
\end{equation}
where $\alpha = 0.9$ is a tolerance margin: spatial positions whose cosine similarity already exceeds~$\alpha$ are excluded from the loss, focusing optimisation on the most misaligned regions.

\section{Experiments}
\label{sec:experiments}
We evaluate our distillation approach on a test subset of StereoLunar containing $1200$ image pairs, covering nadir, pitched, and dynamic geometries, across various altitudes.
Our goal is to assess the trade-off between geometric precision (3D reconstruction and camera pose) and operational efficiency (latency and model size).
Finally, an ablation study on the ViT-Small architecture student isolates the impact of different training strategies.

\subsection{3D reconstruction evaluation}

\myparagraph{3D Reconstruction metrics}
To evaluate the student models, we perform a RANSAC-based optimal similarity transform to align the predicted point clouds with the ground truth (GT), recovering the true metric scale. 
Following the protocol in~\cite{leroy2024grounding}, we report Accuracy (average distance from predicted to closest GT point), Completeness (the reverse), and the Chamfer Distance (the mean of both).
These metrics are reported as Absolute Relative (AbsRel) errors, normalized by the median ground-truth depth to ensure altitude-independence.
To assess terrain safety—critical for lunar landing—we compute the Slope Mean Absolute Error (MAE). Slopes are derived via spatial finite differences on the elevation channel.
This is a classical terrain feature used in landing safety assessments~\cite{Steffes2023}, as it characterizes surface stability.
We also perform a profile-based analysis \cite{Wu2016}, extracting horizontal depth slices to compute statistics along the terrain's cross-sections.

\myparagraph{Camera Pose Estimation metrics} Localization is assessed via the Virtual Correspondence Reprojection Error (VCRE)~\cite{arnold2022mapfree,leroy2024grounding}. 
As in~\cite{arnold2022mapfree}, we report the median VCRE normalised by the image diagonal (\%\,diag), and the Precision at two thresholds: $\text{Prec}@5\%$ and $\text{Prec}@10\%$ of the image diagonal, corresponding to the fraction of pairs with acceptable pose error.

\begin{figure}[t]
    \centering
    \includegraphics[width=1\linewidth]{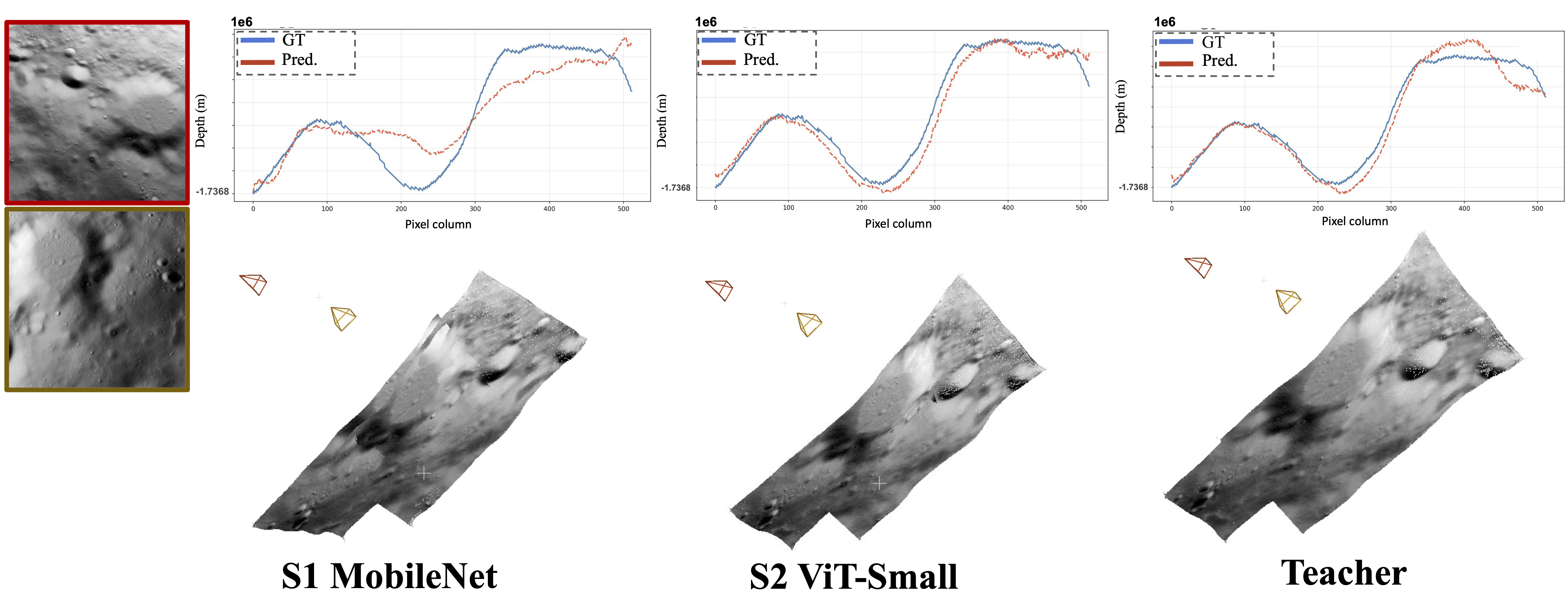}
    \caption{Central profile results for a lunar stereo pair and corresponding 3D scene. Comparison between S2 (best student), S1 (weaker student), and the teacher. S1 exhibits poor cross-view alignment and a central profile poorly correlated with the ground-truth terrain, while S2 produces results closely matching the teacher.}
    \label{fig:example}
\end{figure}

\myparagraph{Results and Discussion}
\cref{tab:quantitative_results} summarizes the performance across the lunar stereo benchmark. Our analysis focuses on the relative performance of the student architectures compared with the teacher model. These quantitative findings are further supported by the qualitative results presented in \cref{fig:teasing}.
We do not include a detailed comparison with generic pretrained models such as MASt3R~\cite{leroy2024grounding} or VGGT~\cite{wang2025vggt}, as these models are not adapted to the lunar domain \cite{grethen2025adaptingstereovisionobjects} and therefore provide limited insight for evaluating the proposed distillation framework.
Focusing first on the geometric reconstruction metrics (Chamfer, Accuracy, and Completeness), we observe a clear correlation between architectural design and reconstruction quality.
Among the student models, S2 achieves the best performance across all three metrics, approaching the teacher despite having more than four times fewer parameters ($154.9$M \vs $688.6$M).
This indicates that a ViT-Small encoder combined with a sufficiently expressive decoder provides a strong balance between capacity and efficiency.
In contrast, S1 replaces the transformer encoder with a lightweight MobileNet backbone while keeping a relatively large decoder. 
Although S1 achieves a slightly smaller parameter count ($135.7$M vs $154.9$M for S2), it results in noticeably worse reconstruction performance, indicating that this compression does not offer a favorable efficiency--accuracy trade-off. This suggests an architectural advantage for self-attention in cross-view reasoning, though pretraining differences between backbones cannot be fully ruled out  (\cref{fig:example} illustrate the poor results of S1 compared to S2 and teacher).
Given these results, we did not explore further reductions of the decoder for S1, as the model already struggles to recover accurate geometry.

The comparison between S3 and S4 further highlights the importance of decoder capacity.
While S4 uses a smaller ViT-Tiny encoder with a relatively large decoder, S3 keeps the same ViT-Small encoder as S2 but uses a reduced decoder ($256$/$4$/$4$).
Despite having fewer parameters ($94.9$M), S3 consistently outperforms S4, suggesting that a stronger encoder with reduced decoder capacity is preferable to a weaker encoder with a larger decoder.
This suggests that maintaining a sufficiently strong encoder while reducing the decoder can be a more effective compression strategy than aggressively shrinking the encoder itself. 
The decoder can be significantly reduced (S3) without
major losses, as long as the encoder retains sufficient capacity to produce view-consistent features.
The VCRE metrics confirm and refine these observations. 
S2 achieves the lowest median reprojection error ($3.56\%$ of image diagonal) and the highest Prec@5\% ($82.8\%$), indicating that its pose estimates are both accurate and consistent.
S3 obtains a nearly identical Prec@10\% ($92.7\%$ \vs $92.1\%$ for S2), suggesting that the two models converge at similar tolerances despite S3 having $40\%$ fewer parameters.
Notably, S1 lags significantly behind on all pose metrics ($\text{Prec}@5\%{=}36.8\%$), reinforcing that the MobileNet encoder fails to learn the geometric structure necessary for reliable relative pose estimation.
Similar trends are observed on terrain-related metrics.
For the elevation profile error (Profile MAE) and slope estimation (Slope MAE), S2 again provides the best trade-off among student models, while S3 remains competitive despite its significantly reduced parameter count.
Overall, these results indicate that moderate decoder reduction combined with a capable transformer encoder yields the most favorable efficiency--accuracy trade-off for lunar stereo reconstruction.

Additional experiments, including the evaluation of alternative encoder architectures (e.g., DUNE) and decoder reduction strategies, as well as further result examples and figures (lighting condition robustness), are provided in the supplementary material.

\begin{table}[t]
\centering
\small
\setlength{\tabcolsep}{4pt}
\begin{tabular}{l c c c c c}
\toprule
\textbf{Metric} & \textbf{S1} & \textbf{S2} & \textbf{S3} & \textbf{S4} & \textbf{Teacher} \\
Encoder & MobileNet & ViT-Small & ViT-Small & ViT-Tiny & ViT-Large \\
Decoder &512/6/4 & 512/6/4 & 256/4/4 & 512/6/4 & 768/12/12 \\
Param(M)   & 135.7 & 154.9 & 94.9 & 138.5 & 688.6\\
\midrule

\multicolumn{6}{l}{} \\
Chamfer $\downarrow$          & 1.02\% & \best{0.53\%} & \second{0.68\%} & 0.74\% & \textbf{0.46\%} \\
Accuracy  $\downarrow$        & 1.09\% & \best{0.56\%} & \second{0.70\%} & 0.80\% & \textbf{0.46\%} \\
Completeness $\downarrow$     & 0.95\% & \best{0.51\%} & \second{0.65\%} & 0.68\% & \textbf{0.46\%} \\
\midrule
\multicolumn{6}{l}{\small\textit{VCRE~\cite{arnold2022mapfree}}} \\[2pt]
    Med.\ (\% diag) $\downarrow$  & 8.75 & \best{3.56} & \second{3.57} & 4.04 & \textbf{1.15} \\
Prec@5\%  $\uparrow$          & 36.8 & \best{82.8} & \second{79.2} & 76.5 & \textbf{98.6} \\
Prec@10\% $\uparrow$          & 69.2 & \second{92.1}        & \best{92.7}   &   91.7 & \textbf{99.3} \\

\midrule
\multicolumn{6}{l}{} \\

Profile MAE (m) $\downarrow$ & 161.85 & \best{75.69} & \second{85.82} & 93.41& \textbf{53.63} \\
Slope MAE ($^\circ$) $\downarrow$ & 8.37 & \best{7.59} & \second{7.97} & 8.03 & \textbf{7.22} \\

\bottomrule
\end{tabular}

\caption{Quantitative comparison of student models against the Teacher on the test dataset. \emph{3D reconstruction} is measured via Chamfer distance, Accuracy, and Completeness. 
\emph{Pose quality} is assessed via VCRE~\cite{arnold2022mapfree}: median reprojection error (\%\,diag) and Precision at \SI{5}{\percent} and \SI{10}{\percent} of the image diagonal. \emph{Terrain} metrics measure depth profile fidelity (Profile MAE) and slope estimation (Slope MAE). \best{Green} = best student, \second{orange} = second best, \textbf{bold} = Teacher (baseline).}

\label{tab:quantitative_results}
\end{table}

\subsection{Inference Efficiency}
\begin{table}[t]
\centering
\small
\setlength{\tabcolsep}{4pt}
\begin{tabular}{l r r r r r}
\toprule
Model & Params & N=2 & N=32 & N=64 & N=128 \\
\midrule
Teacher              & 688.6M & 0.07s & 2.04s & 4.17s & 8.36s \\
\midrule
S1          & 135.7M & 0.04s & 0.91s & 1.84s & 3.69s \\
S2             & 154.9M & 0.04s & 0.98s & 1.99s & 4.01s \\
S3     &  94.9M & 0.04s & 0.86s & 1.73s & 3.47s \\
S4           & 138.5M & 0.03s & 0.87s & 1.77s & 3.55s \\
\bottomrule
\end{tabular}
\caption{Inference time and model size. $N$ denotes the number of image pairs. All students achieve $\sim\!2\times$ speedup over the teacher with $4$--$7\times$ fewer parameters. Measured on a single NVIDIA A100 GPU.}
\label{tab:inference}
\end{table}

\Cref{tab:inference} reports inference time and parameter count for each architecture. All students achieve $\approx2\times$ speedup over the teacher, with inference scaling linearly with pair count. S3 is the most compact model ($7.3\times$ smaller, $2.4\times$ faster at $N\!=\!128$), while S2 offers the best accuracy-efficiency trade-off ($4.4\times$ fewer parameters, $2.1\times$ faster, 15\% Chamfer drop). At small batch sizes ($N=2$), students run in \qtyrange{0.03}{0.04}{\second} vs.\ \SI{0.07}{\second} for the teacher, which is particularly relevant for deployment on resource-constrained platforms

\subsection{Training Ablation}
In the following, we present the ablation study on S2, as it appears to offer the best combination of performance, size, and inference time.
We conduct a series of experiments on the S2 (ViT-Small) student to identify the best distillation configuration, varying the supervision source, the encoder training strategy, and the loss composition (\cref{tab:ablation_s2}).
All experiments share the same architecture and training data; only one factor is changed at a time. 

\myparagraph{Distillation \vs ground-truth supervision.}
Replacing the teacher's pseudo-labels with ground-truth depth and pose (Exp.~E) yields slightly worse results (Chamfer: \SI{0.65}{\percent} \vs \SI{0.53}{\percent}). 
This validates the core premise of our approach: The full pipeline (Exp.~A, knowledge distillation (KD) + SVD + feature alignment) nonetheless outperforms Exp.~E by a large margin, confirming that the teacher's dense per-pixel predictions, combined with structured initialization and representation alignment, provide a
richer supervisory signal than ground-truth labels alone.
By distilling from the teacher, the student benefits from implicit geometric priors learned during the teacher's own training, effectively transferring knowledge that standard supervised losses cannot capture. Importantly, this result also shows that the proposed training scheme remains applicable even when accurate ground-truth annotations are unavailable, a common situation in many real-world 3D reconstruction scenarios.

\myparagraph{Feature alignment loss.}
Removing the intermediate feature alignment loss (Exp.~D) increases the Chamfer from \SI{0.53}{\percent} to \SI{0.67}{\percent}. 
Supervising the encoder's internal representations, not just its final output, is therefore important for effective distillation across the large capacity gap between teacher and student.

\myparagraph{SVD initialisation.}
Initialising the student decoder from a truncated SVD projection of the teacher's weights (Exp.~A vs.\ Exp.~B) reduces the Chamfer from \SI{0.68}{\percent} to \SI{0.53}{\percent}. 
Without this initialisation, the student must learn the cross-attention and regression layers from scratch, resulting in a significantly worse starting point for optimisation.
The SVD projection preserves the dominant modes of the teacher's decoder weights, giving the student a warm start that accelerates convergence and improves final performance.

\myparagraph{Encoder freezing.}
Freezing the DINOv2 backbone (Exp.~C) leads to the sharpest performance drop (Chamfer: \SI{0.87}{\percent}), indicating that the pretrained features, while providing a strong initialisation, must be adapted to the lunar domain during fine-tuning.

\begin{table}[t]
\centering
\caption{Ablation study on S2 (DINOv2 ViT-Small encoder). Each column modifies one factor relative to our best configuration (Exp.~A) except for Exp. E which serves as a fully-supervised baseline. AbsRel metrics (lower is better).}
\label{tab:ablation_s2}
\setlength{\tabcolsep}{3.5pt}
\begin{tabular}{l c c c c c c}
\toprule
 & \textit{Teacher} & \textbf{A}  & \textbf{B}  & \textbf{C}  & \textbf{D}  & \textbf{E}  \\
\midrule
KD             & --  & \cmark & \cmark & \cmark & \cmark & \xmark \\
Feature loss   & --  & \cmark & \cmark & \cmark & \xmark & \xmark \\
SVD init       & --  & \cmark & \xmark & \cmark & \cmark & \xmark \\
Unfrozen enc.  & --  & \cmark & \cmark & \xmark & \cmark & \cmark \\
\midrule
Chamfer$\downarrow$      & \textit{0.46\%} & \textbf{0.53\%} & 0.68\% & 0.87\% & 0.67\% & 0.65\% \\
Accuracy$\downarrow$     & \textit{0.46\%} & \textbf{0.56\%} & 0.69\% & 0.85\% & 0.70\% & 0.69\% \\
Completeness$\downarrow$ & \textit{0.46\%} & \textbf{0.51\%} & 0.67\% & 0.89\% & 0.63\% & 0.61\% \\
\bottomrule
\end{tabular}
\end{table}

\section{Conclusions}
\label{sec:conclusions}

We presented a systematic distillation study for compressing stereo 3D 
reconstruction models, using lunar imagery as a challenging testbed.
Our results highlight five transferable lessons.
First, regardless of whether the gap stems from architecture or pretraining, a well-pretrained ViT encoder remains the safer practical choice over a CNN backbone.
Second, when further compression is required, reducing decoder capacity is preferable to reducing encoder capacity: moderate decoder reduction combined with a capable transformer encoder yields the most favorable efficiency--accuracy trade-off.
Third, teacher pseudo-labels provide a richer supervisory signal than sparse ground-truth annotations, making the approach applicable to domains where labels are scarce.
Fourth, feature-level alignment between encoder representations is necessary to close the capacity gap, even when teacher and student have mismatched dimensions.
Fifth, structured SVD-based initialization of the student decoder from the teacher's weights improves optimization stability and final reconstruction accuracy compared to random initialization.
While our analysis focuses on the lunar domain, the results suggest that this distillation strategy --- domain-adapted teacher, pseudo-GT supervision, representation alignment, and structured initialization--- provides a methodological template potentially applicable to other resource-constrained geometric tasks domain where a large pretrained model is available.
\section*{Acknowledgments}
This work was supported by the French Agence Nationale de la Recherche (ANR, “Investissements d’avenir”, ANR-21-ESRE-0051) and the European Space Agency (ESA, contract 4000140461/23/NL/GLC/my).

\clearpage
\appendix
\setcounter{section}{0}
\renewcommand{\thesection}{\Alph{section}}

\begin{center}
    {\LARGE \textbf{Appendix}}
\end{center}
\vspace{1em}
In this supplementary material, we present additional results that complement the study detailed in the main paper. \cref{sec:details} 
provides additional details on the distillation setup.
\cref{sec:lighting_robust} provides further evaluations of robustness under varying lighting conditions. \cref{sec:training_ablation} presents qualitative results for the ablation study across different training setups. 
\cref{sec:new_students} introduces an additional comparison using the DUNE encoder. Finally, \cref{sec:svd} provides a comprehensive justification for the SVD-based initialization, detailing both our design choices and implementation.
\section{Implementation details}
\label{sec:details}
The teacher model is the MASt3R model fine-tuned in StereoLunar \cite{grethen2025adaptingstereovisionobjects} for 25 epochs using a learning rate of $3\times10^{-5}$ and a cosine learning rate schedule. During distillation, we set the confidence weighting parameter of the geometric loss to $\beta = 0.2$. All images are processed at a resolution of $512 \times 384$, corresponding to the standard DUSt3R crop. 

\section{Additional qualitative results: lighting condition robustness}
\label{sec:lighting_robust}

\begin{figure}
    \centering
    \includegraphics[width=1\linewidth]{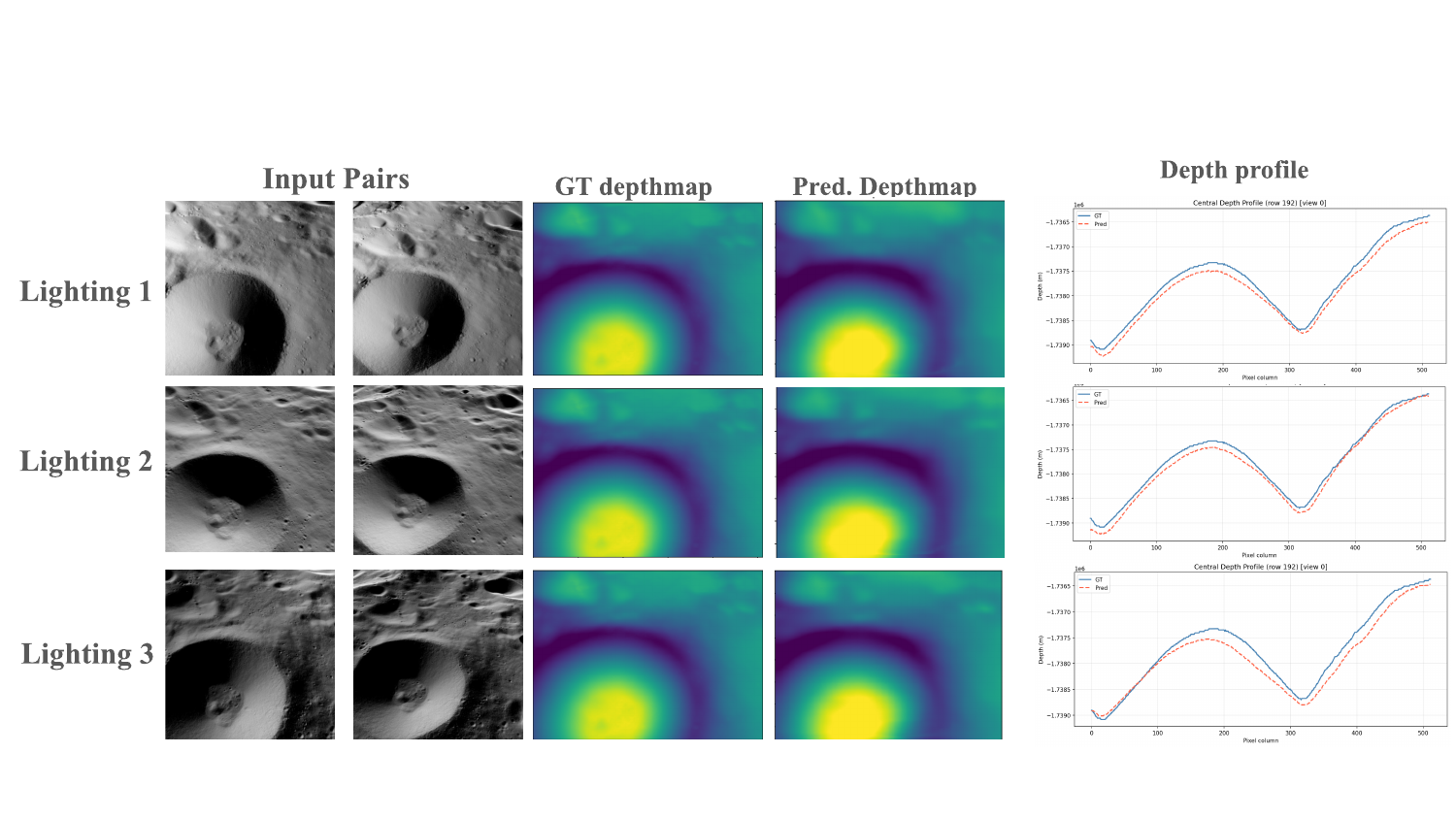}
    \caption{Robustness to lighting variations on a challenging lunar crater scene. The same scene is evaluated, using S2 student, under three different illumination conditions from the test set. The first column shows the input stereo pairs under the three lighting setups, the second column shows the ground-truth (GT) depth maps, the third column shows the predicted depth maps, and the fourth column shows the central depth profiles comparing GT (solid blue) and prediction (dashed red). }
    \label{fig:light}
\end{figure}

To further assess the robustness of the best compromise  student model (S2 Small-ViT encoder), we evaluate it on the same stereo pair captured under three distinct lighting conditions drawn from the test set (\cref{fig:light}). 
This experiment targets a particularly challenging scenario: lunar crater surfaces, where the photometric ambiguity between convex and concave shapes is well-known — a crater can appear either as a bowl or a mound depending on the direction of illumination~\cite{Belhumeur1999}. 
Despite this perceptual ambiguity, our student model consistently recovers the correct concave structure under all lighting conditions, without degrading prediction quality.
The central depth profiles confirm this robustness quantitatively: the predicted profiles (dashed red) remain tightly aligned with the ground truth (solid blue) in all three cases, demonstrating that the model has learned geometry-driven features rather than relying on illumination cues.

\section{Qualitative training ablation analysis}
\label{sec:training_ablation}

\begin{figure}
    \centering
    \includegraphics[width=1\linewidth]{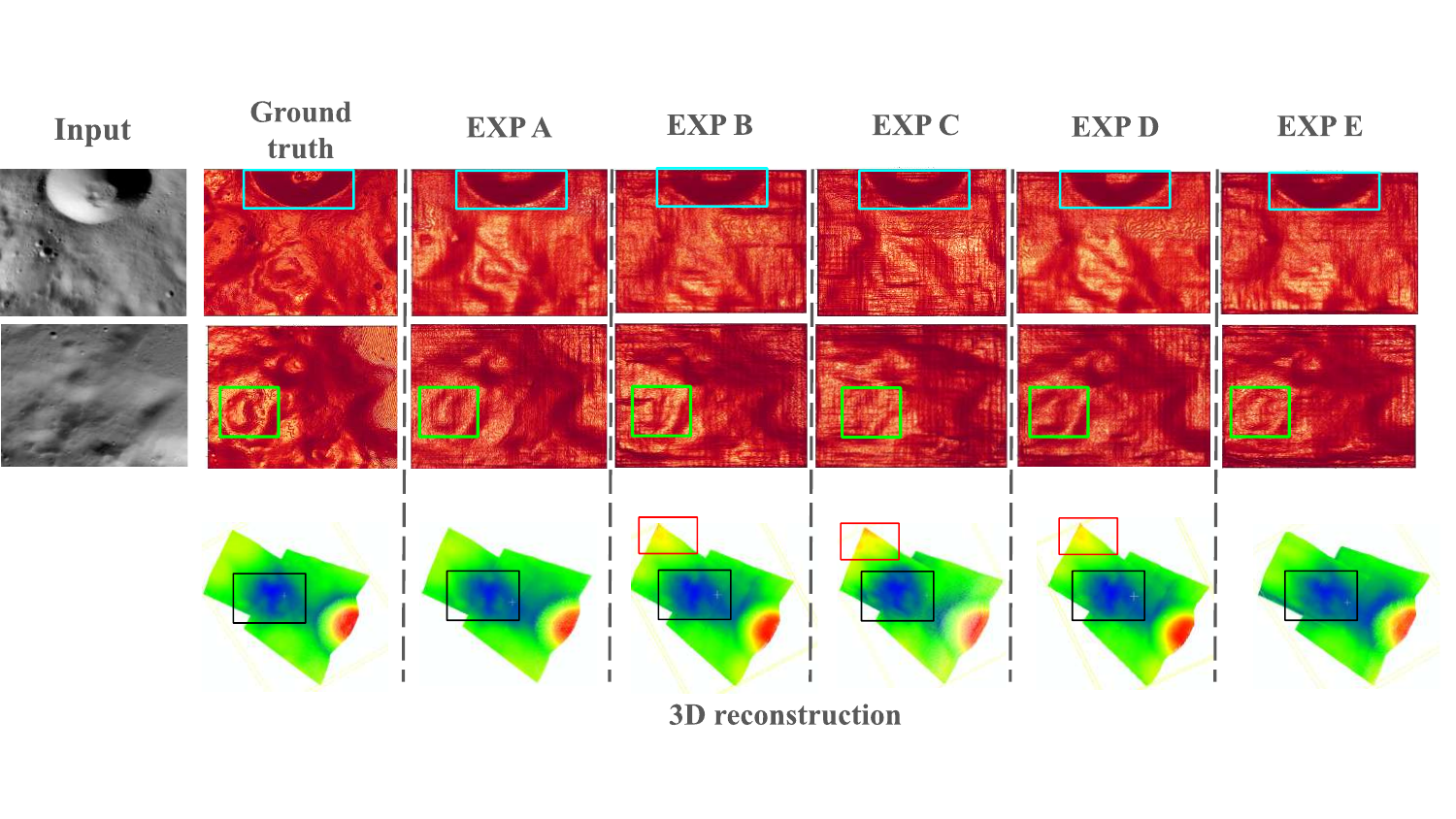}
    \caption{Qualitative ablation results on one stereo pair with a challenging camera baseline for the S2 student. 
    Rows 1--2 show the predicted surface slope maps for each view, while Row 3 shows the reconstructed 3D scene (colors indicate depth, with blue corresponding to closer points and red to farther points). 
    Colored boxes highlight regions of interest discussed in the text. 
    Exp.~A corresponds to the full model, while each subsequent configuration removes one component as detailed in Tab. 3 of the main paper.}
    \label{fig:ablation_visual}

\end{figure}

\cref{fig:ablation_visual} provides qualitative illustrations corresponding to the quantitative trends reported in the main paper, showing the results of the different experiments conducted in the ablation study of the training pipeline on the S2 (ViT-Small) student.
Colored bounding boxes highlight discriminative regions across configurations.
Exp.~A, which combines all components (feat loss and SVD), produces the sharpest slope maps and the most geometrically consistent 3D reconstruction, with well-aligned point clouds visible in the \textcolor{black}{black} box. 
Exp.~C, where the encoder is frozen, yields the most severe degradation: both the \textcolor{cyan}{cyan} and \textcolor{green}{green} boxes reveal heavily noisy, distorted slope maps with lost surface structure, and the 3D reconstruction shows misaligned point clouds, consistent with its largest metric gap.
Exp.~B, lacking SVD initialization, produces globally reasonable slope maps but introduces a spurious planar artifact in the 3D reconstruction (visible in the \textcolor{red}{red} box, also visible in Exp.~C and Exp.~D), reflecting incomplete cross-view alignment. 
Exp.~D and Exp.~E exhibit milder degradation in the \textcolor{green}{green} boxes compared to Exp.~C, with slightly blurred crater boundaries but overall preserved geometry, consistent with their more moderate quantitative drops.
Overall, the qualitative results confirm that the unfrozen encoder is the most critical component, while SVD initialization, feature alignment, and knowledge distillation each contribute incrementally to the final reconstruction quality.

\section{Results on new students architectures}
\label{sec:new_students}

\begin{table}[t]
\centering
\small
\setlength{\tabcolsep}{4pt}
\begin{tabular}{l c c c c }
\toprule
\textbf{Metric} & \textbf{Dune S5} & \textbf{Dune S6} & \textbf{Dune S7} & \textbf{S2}  \\
Encoder & Dune Small & Dune Small & Dune Small & ViT-Small (DINO) \\
Decoder & 384/6/6 & 384/6/6 & 512/6/4 & 512/6/4 \\
Param(M)   & 123.5 & 123.5 & 154.8 & 154.9 \\
\midrule

\multicolumn{5}{l}{} \\
Chamfer $\downarrow$          & 1.29\% & 1.21\% & \second{1.05\%} & \best{0.90}\%  \\
Accuracy  $\downarrow$        & 1.38\% & 1.29\% & \second{1.14\%} & \best{0.98}\% \\
Completeness $\downarrow$     & 1.21\% & 1.13\% & \second{0.97\%} & \best{0.82}\%  \\
\midrule
\multicolumn{5}{l}{\small\textit{VCRE~\cite{arnold2022mapfree}}} \\[2pt]
Med.\ (\% diag) $\downarrow$  & 27.30  & \second{ 16.39} &  17.29 & \best{9.38}  \\
Prec@5\%  $\uparrow$          &  27.8 & 29.5 & \second{29.8} & \best{45.9}  \\
Prec@10\% $\uparrow$          & 37.4 & 43.7       & \second{44.2}   & \best{68.3 } \\

\midrule
\multicolumn{5}{l}{} \\

Profile MAE (m) $\downarrow$ & 178.79  & 140.16 & \second{135.71} & \best{135.37 }\\
Slope MAE ($^\circ$) $\downarrow$ &  8.97 & 8.74 & \second{8.43 } & \best{8.37} \\

\bottomrule
\end{tabular}

\caption{Quantitative comparison of student models at epoch 10 (Dune and Dino encoders). \emph{3D reconstruction} is measured via Chamfer distance, Accuracy, and Completeness. 
\emph{Pose quality} is assessed via VCRE~\cite{arnold2022mapfree}: median reprojection error (\%\,diag) and Precision at \SI{5}{\percent} and \SI{10}{\percent} of the image diagonal. \emph{Terrain} metrics measure depth profile fidelity (Profile MAE) and slope estimation (Slope MAE). \best{Green} = best student, \second{orange} = second best.}
\label{tab:dune_result}
\end{table}

Inspired by Distill3R~\cite{leblanc2026distill3r}, which distills Fast3R into a DUNE ViT-S student~\cite{sariyildiz2025dune}, we investigate whether this encoder -- pretrained via multi-teacher distillation (DINOv2, MASt3R, Multi-HMR) -- can be effective in our framework. 
We define three configurations: \textbf{Dune~S5} inspired by the Distill3R setup (DUNE encoder, $384$-dim/$6$-layer/$6$-head decoder, geometric loss only);
\textbf{Dune~S6} adds our training pipeline (SVD init + feature alignment); and \textbf{Dune~S7} further adopts our reference decoder ($512$-dim/$6$-layer/$4$-head), matching S2's architecture to isolate the encoder effect at equal parameter budget ($\approx155$\,M).

\cref{tab:dune_result} confirms a clear progression. 
Adding our training strategy (S5$\to$S6) reduces Chamfer from \SI{1.29}{\percent} to \SI{1.21}{\percent} and reduces the median VCRE by 40\%
($27.30\to16.39$). 
Scaling the decoder (S6$\to$S7) yields further gains (Chamfer 1.05\%, Profile~MAE \SI{135.71}{\metre}), at the cost of $\approx31$\,M additional parameters (\SI{+25}{\percent}).

However, a gap persists between S7 (DUNE) and S2 (DINOv2), which share the same ViT-S/14 architecture and decoder. 
The difference is moderate in depth---Profile~MAE is nearly identical---but pronounced on pose estimation, where S2 improves precision by $\sim\!54\%$ over S7 (Prec@5\%: \SI{45.9}{\percent} \vs \SI{29.8}{\percent}).
Since the architecture is identical, this gap stems from pre-training: DINOv2's general-purpose self-supervised features provide a neutral starting point easily steered toward MASt3R by our alignment loss, whereas DUNE's features, shaped by three heterogeneous teachers, represent a multi-task compromise that may resist realignment -- a form of negative transfer most visible on correspondence-sensitive tasks like pose estimation.

These results suggest that our SVD + feature alignment pipeline generalises across encoder types, but that the encoder's pre-training compatibility with the teacher remains an important factor for distillation quality.
Due to the heavy computational time required, these preliminary experiments were conducted over $10$ epochs (including S2). 
Further training would be needed to confirm the observed trends.
\section{SVD initialization}
\label{sec:svd}
\noindent\textbf{\textit{SVD choice.}} 
By the Eckart-Young theorem, SVD truncation yields the \emph{optimal} rank-$d_s$ approximation of the teacher weights under Frobenius norm — a principled choice over random init (ablated in the paper) or direct truncation, which assumes importance is axis-aligned.

\noindent\textit{\textbf{Layer mapping sensitivity.}}
This ablation was conducted as a \emph{preliminary} experiment — before launching the full training runs — specifically to validate the mapping choice early and avoid committing compute to an arbitrary design decision.
We trained $25$-epoch students on a subset of the lunar data and evaluated on a held-out subset.
All three strategies yield similar results (mean Chamfer absrel: uniform \SI{0.87}{\percent}, last-$k$ \SI{0.87}{\percent}, first-$k$ \SI{0.85}{\percent}), with differences under $0.02$ pp.
No strategy consistently dominates, and so the averaged gap is within noise.
Uniform is also the standard choice for weight-based init in transformer distillation \cite{sanh2020distilbertdistilledversionbert}, and layer-selection insensitivity has been empirically validated in prior work \cite{yu2025revisiting}; our SVD step generalizes this to the cross-dimension case.
\section{Supplementary material conclusions}
\label{sec:conclusions}
First, we show that the proposed student model (S2) remains robust 
under challenging light variations, correctly recovering surface 
geometry despite photometric ambiguities. Second, qualitative results 
on the S2 training ablation confirm the quantitative trends from the 
main paper, highlighting the unfrozen encoder as the most critical 
component, with SVD initialization, feature alignment, and knowledge 
distillation each contributing incrementally. Finally, experiments 
with the Dune encoder demonstrate that our training pipeline 
generalises across encoder architectures, while highlighting the 
importance of encoder-teacher pre-training compatibility.
\bibliographystyle{splncs04}
\bibliography{main}

@String(IJCV  = {Int. J. Comput. Vis.})

@String(CVPR  = {IEEE Conf. Comput. Vis. Pattern Recog.})

@String(ICCV  = {Int. Conf. Comput. Vis.})

@String(ECCV  = {Eur. Conf. Comput. Vis.})

@String(ICLR  = {Int. Conf. Learn. Represent.})

@STRING{iccvw = "{IEEE International Conference on Computer Vision Workshops}"}

@String(IJCV  = {IJCV})

@String(CVPR  = {CVPR})

@String(ICCV  = {ICCV})

@String(ECCV  = {ECCV})

@String(ICLR  = {ICLR})

@String(IJCV  = {International Journal of Computer Vision})

@inproceedings{grethen2025adaptingstereovisionobjects,
  title={Adapting Stereo Vision From Objects To 3D Lunar Surface Reconstruction with the StereoLunar Dataset},
  author={Grethen, Cl{\'e}mentine and Gasparini, Simone and Morin, G{\'e}raldine and Lebreton, Jeremy and Marti, Lucas and Sanchez-Gestido, Manuel},
  booktitle={iccvw},
  pages={3751--3760},
  year={2025}
}

@inproceedings{leroy2024grounding,
    author = {Leroy, Vincent and Cabon, Yohann and Revaud, Jerome},
    title = {Grounding Image Matching in 3D with {MASt3R}},
    year = {2024},
    XXisbn = {978-3-031-73219-5},
    XXpublisher = {Springer-Verlag},
    XXaddress = {Berlin, Heidelberg},
    XXurl = {https://doi.org/10.1007/978-3-031-73220-1_5},
    XXdoi = {10.1007/978-3-031-73220-1_5},
    booktitle = eccv,
    XXpages = {71–91},
    numpages = {21},
    XXlocation = {Milan, Italy}
}

@article{wang2023dust3r,
  title={Dust3r: Geometric 3d vision made easy},
  author={Wang, Shuzhe and Leroy, Vincent and Cabon, Yohann and Chidlovskii, Boris and Revaud, Jerome},
  journal={arXiv preprint arXiv:2310.02328},
  year={2023}
}

@article{1979,
  title = {The interpretation of structure from motion},
  volume = {203},
  ISSN = {2053-9193},
  XXurl = {http://dx.doi.org/10.1098/rspb.1979.0006},
  DOI = {10.1098/rspb.1979.0006},
  number = {1153},
  journal = {Proceedings of the Royal Society of London. Series B. Biological Sciences},
  publisher = {The Royal Society},
  author = {Ullman,  S.},
  year = {1979},
  month = jan,
  pages = {405–426}
}

@inproceedings{wang2025vggt,
  title={VGGT: Visual Geometry Grounded Transformer},
  author={Wang, Jianyuan and Chen, Minghao and Karaev, Nikita and Vedaldi, Andrea and Rupprecht, Christian and Novotny, David},
  booktitle={Proceedings of the IEEE/CVF Conference on Computer Vision and Pattern Recognition},
  year={2025}
}

@misc{dosovitskiy2021imageworth16x16words,
      title={An Image is Worth 16x16 Words: Transformers for Image Recognition at Scale}, 
      author={Alexey Dosovitskiy and Lucas Beyer and Alexander Kolesnikov and Dirk Weissenborn and Xiaohua Zhai and Thomas Unterthiner and Mostafa Dehghani and Matthias Minderer and Georg Heigold and Sylvain Gelly and Jakob Uszkoreit and Neil Houlsby},
      year={2021},
      eprint={2010.11929},
      archivePrefix={arXiv},
      primaryClass={cs.CV},
      url={https://arxiv.org/abs/2010.11929}, 
}

@inproceedings{wu2022tinyvit,
  title={Tinyvit: Fast pretraining distillation for small vision transformers},
  author={Wu, Kan and Zhang, Jinnian and Peng, Houwen and Liu, Mengchen and Xiao, Bin and Fu, Jianlong and Yuan, Lu},
  booktitle=eccv,
  pages={68--85},
  year={2022},
  organization={Springer}
}

@inbook{Getchius2024,
  title = {{Hazard Detection and Avoidance for the Nova-C Lander}},
  ISBN = {9783031519284},
  ISSN = {2731-0884},
   DOI = {10.1007/978-3-031-51928-4_53},
  booktitle = {Proceedings of the 44th Annual American Astronautical Society Guidance,  Navigation,  and Control Conference,  2022},
  publisher = {Springer International Publishing},
  author = {Getchius,  Joel and Renshaw,  Devin and Posada,  Daniel and Henderson,  Troy and Hong,  Lillian and Ge,  Shen and Molina,  Giovanni},
  year = {2024},
  pages = {921–943}
}

@misc{kumar2024moonmetasynclunarimageregistration,
      title={MoonMetaSync: Lunar Image Registration Analysis}, 
      author={Ashutosh Kumar and Sarthak Kaushal and Shiv Vignesh Murthy},
      year={2024},
      eprint={2410.11118},
      archivePrefix={arXiv},
      primaryClass={cs.CV},
      url={https://arxiv.org/abs/2410.11118}, 
}

@misc{leblanc2026distill3r,
  title={Distill3R: A Pipeline for Democratizing 3D Foundation Models on Commodity Hardware}, 
  author={Brandon Leblanc and Charalambos Poullis},
  year={2026},
  eprint={2602.00865},
  archivePrefix={arXiv},
  primaryClass={cs.CV},
  XXurl={https://arxiv.org/abs/2602.00865}, 
}

@article{Gou2021,
  title = {Knowledge Distillation: A Survey},
  volume = {129},
  ISSN = {1573-1405},
  XXurl = {http://dx.doi.org/10.1007/s11263-021-01453-z},
  DOI = {10.1007/s11263-021-01453-z},
  number = {6},
  journal = IJCV,
  publisher = {Springer Science and Business Media LLC},
  author = {Gou,  Jianping and Yu,  Baosheng and Maybank,  Stephen J. and Tao,  Dacheng},
  year = {2021},
  month = mar,
  pages = {1789–1819}
}

@misc{sanh2020distilbertdistilledversionbert,
      title={DistilBERT, a distilled version of BERT: smaller, faster, cheaper and lighter}, 
      author={Victor Sanh and Lysandre Debut and Julien Chaumond and Thomas Wolf},
      year={2020},
      eprint={1910.01108},
      archivePrefix={arXiv},
      primaryClass={cs.CL},
      url={https://arxiv.org/abs/1910.01108}, 
}

@misc{xu2024surveyknowledgedistillationlarge,
      title={A Survey on Knowledge Distillation of Large Language Models}, 
      author={Xiaohan Xu and Ming Li and Chongyang Tao and Tao Shen and Reynold Cheng and Jinyang Li and Can Xu and Dacheng Tao and Tianyi Zhou},
      year={2024},
      eprint={2402.13116},
      archivePrefix={arXiv},
      primaryClass={cs.CL},
      url={https://arxiv.org/abs/2402.13116}, 
}

@inproceedings{touvron2021training,
  title={Training data-efficient image transformers \& distillation through attention},
  author={Touvron, Hugo and Cord, Matthieu and Douze, Matthijs and Massa, Francisco and Sablayrolles, Alexandre and J{\'e}gou, Herv{\'e}},
  booktitle={International conference on machine learning},
  pages={10347--10357},
  year={2021},
  organization={PMLR}
}

@article{hinton2015distilling,
  title={Distilling the knowledge in a neural network},
  author={Hinton, Geoffrey and Vinyals, Oriol and Dean, Jeff},
  journal={arXiv preprint arXiv:1503.02531},
  year={2015}

}

@inproceedings{yu2025revisiting,
  title={Revisiting intermediate-layer matching in knowledge distillation: Layer-selection strategy doesn’t matter (much)},
  author={Yu, Zony and Wen, Yuqiao and Mou, Lili},
  booktitle={Proceedings of the 14th International Joint Conference on Natural Language Processing and the 4th Conference of the Asia-Pacific Chapter of the Association for Computational Linguistics},
  pages={1686--1694},
  year={2025}
}

@article{streamVGGT,
      title={Streaming 4D Visual Geometry Transformer}, 
      author={Dong Zhuo and Wenzhao Zheng and Jiahe Guo and Yuqi Wu and Jie Zhou and Jiwen Lu},
      journal={arXiv preprint arXiv:2507.11539},
      year={2025}
}

@misc{guo2025gladgeometriclatentdistillation,
      title={GLaD: Geometric Latent Distillation for Vision-Language-Action Models}, 
      author={Minghao Guo and Meng Cao and Jiachen Tao and Rongtao Xu and Yan Yan and Xiaodan Liang and Ivan Laptev and Xiaojun Chang},
      year={2025},
      eprint={2512.09619},
      archivePrefix={arXiv},
      primaryClass={cs.RO},
      url={https://arxiv.org/abs/2512.09619}, 
}

@misc{dutt2026multiview,
      title={Multi-View 3D Reconstruction using Knowledge Distillation}, 
      author={Aditya Dutt and Ishikaa Lunawat and Manpreet Kaur},
      year={2026},
      eprint={2412.02039},
      archivePrefix={arXiv},
      primaryClass={cs.CV},
      url={https://arxiv.org/abs/2412.02039}, 
}

@article{weinzaepfel2022croco,
  title={Croco: Self-supervised pre-training for 3d vision tasks by cross-view completion},
  author={Weinzaepfel, Philippe and Leroy, Vincent and Lucas, Thomas and Br{\'e}gier, Romain and Cabon, Yohann and Arora, Vaibhav and Antsfeld, Leonid and Chidlovskii, Boris and Csurka, Gabriela and Revaud, J{\'e}r{\^o}me},
  journal={Advances in Neural Information Processing Systems},
  volume={35},
  pages={3502--3516},
  year={2022}
}

@InProceedings{Howard_2019_ICCV,
author = {Howard, Andrew and Sandler, Mark and Chu, Grace and Chen, Liang-Chieh and Chen, Bo and Tan, Mingxing and Wang, Weijun and Zhu, Yukun and Pang, Ruoming and Vasudevan, Vijay and Le, Quoc V. and Adam, Hartwig},
title = {Searching for MobileNetV3},
booktitle = {Proceedings of the IEEE/CVF International Conference on Computer Vision (ICCV)},
month = {October},
year = {2019}
}

@article{oquab2023dinov2,
  title={Dinov2: Learning robust visual features without supervision},
  author={Oquab, Maxime and Darcet, Timoth{\'e}e and Moutakanni, Th{\'e}o and Vo, Huy and Szafraniec, Marc and Khalidov, Vasil and Fernandez, Pierre and Haziza, Daniel and Massa, Francisco and El-Nouby, Alaaeldin and others},
  journal={arXiv preprint arXiv:2304.07193},
  year={2023}
}

@article{vaswani2017attention,
  title={Attention is all you need},
  author={Vaswani, Ashish and Shazeer, Noam and Parmar, Niki and Uszkoreit, Jakob and Jones, Llion and Gomez, Aidan N and Kaiser, {\L}ukasz and Polosukhin, Illia},
  journal={Advances in neural information processing systems},
  volume={30},
  year={2017}
}

@article{sy2024lillama,
  title={Lillama: Large language models compression via low-rank feature distillation},
  author={Sy, Yaya and Cerisara, Christophe and Illina, Irina},
  journal={arXiv preprint arXiv:2412.16719},
  year={2024}
}

@article{eckart1936approximation,
  title={The approximation of one matrix by another of lower rank},
  author={Eckart, Carl and Young, Gale},
  journal={Psychometrika},
  volume={1},
  number={3},
  pages={211--218},
  year={1936},
  publisher={Springer-Verlag}
}

@article{hu2022lora,
  title={Lora: Low-rank adaptation of large language models.},
  author={Hu, Edward J and Shen, Yelong and Wallis, Phillip and Allen-Zhu, Zeyuan and Li, Yuanzhi and Wang, Shean and Wang, Liang and Chen, Weizhu and others},
  journal={Iclr},
  volume={1},
  number={2},
  pages={3},
  year={2022}
}

@article{romero2015fitnets,
  title={Fitnets: Hints for thin deep nets. In International conference on learning representations},
  author={Romero, A and Ballas, N and Kahou, SE and Chassang, A and Gatta, C and Bengio, Y},
  year={2015}
}

@inproceedings{depthanything,
  title={Depth Anything: Unleashing the Power of Large-Scale Unlabeled Data},
  author={Yang, Lihe and Kang, Bingyi and Huang, Zilong and Xu, Xiaogang and Feng, Jiashi and Zhao, Hengshuang},
  booktitle={CVPR},
  year={2024}
}

@inproceedings{Steffes2023,
  title = {Hazard Boresight Relative Navigation for Safe Lunar Landing},
  XXurl = {http://dx.doi.org/10.2514/6.2023-0691},
  DOI = {10.2514/6.2023-0691},
  booktitle = {AIAA SCITECH 2023 Forum},
  publisher = {American Institute of Aeronautics and Astronautics},
  author = {Steffes,  Stephen R. and DeTrempe,  Paul and Barton,  Gregory and Woffinden,  David},
  year = {2023},
  month = jan 
}

@article{Wu2016,
  title = {SHAPE AND ALBEDO FROM SHADING ({SAfS}) FOR PIXEL-LEVEL DEM GENERATION FROM MONOCULAR IMAGES CONSTRAINED BY LOW-RESOLUTION DEM},
  volume = {XLI-B4},
  ISSN = {2194-9034},
  XXurl = {http://dx.doi.org/10.5194/isprs-archives-XLI-B4-521-2016},
  DOI = {10.5194/isprs-archives-xli-b4-521-2016},
  journal = {The International Archives of the Photogrammetry,  Remote Sensing and Spatial Information Sciences},
  publisher = {Copernicus GmbH},
  author = {Wu,  Bo and Liu,  Wai Chung and Grumpe,  Arne and W\"{o}hler,  Christian},
  year = {2016},
  month = jun,
  pages = {521–527}
}

@inproceedings{arnold2022mapfree,
      title={Map-free Visual Relocalization: Metric Pose Relative to a Single Image},
      author={Arnold, Eduardo and Wynn, Jamie and Vicente, Sara and Garcia-Hernando, Guillermo and Monszpart, {\'{A}}ron and Prisacariu, Victor Adrian and Turmukhambetov, Daniyar and Brachmann, Eric},
      booktitle={ECCV},
      year={2022},
    }

@misc{vuong2025improvingroboticmanipulationefficient,
      title={Improving Robotic Manipulation with Efficient Geometry-Aware Vision Encoder}, 
      author={An Dinh Vuong and Minh Nhat Vu and Ian Reid},
      year={2025},
      eprint={2509.15880},
      archivePrefix={arXiv},
      primaryClass={cs.RO},
      url={https://arxiv.org/abs/2509.15880}, 
}

@inproceedings{sariyildiz2025dune,
  title={Dune: Distilling a universal encoder from heterogeneous 2d and 3d teachers},
  author={Sar{\i}y{\i}ld{\i}z, Mert B{\"u}lent and Weinzaepfel, Philippe and Lucas, Thomas and De Jorge, Pau and Larlus, Diane and Kalantidis, Yannis},
  booktitle={Proceedings of the Computer Vision and Pattern Recognition Conference},
  pages={30084--30094},
  year={2025}
}

@inproceedings{baradel2024multi,
  title={Multi-hmr: Multi-person whole-body human mesh recovery in a single shot},
  author={Baradel, Fabien and Armando, Matthieu and Galaaoui, Salma and Br{\'e}gier, Romain and Weinzaepfel, Philippe and Rogez, Gr{\'e}gory and Lucas, Thomas},
  booktitle={European Conference on Computer Vision},
  pages={202--218},
  year={2024},
  organization={Springer}
}

@Book{Hartley2004,
    author = "Hartley, R.~I. and Zisserman, A.",
    title = "Multiple View Geometry in Computer Vision",
    edition = "Second",
    year = "2004",
    publisher = "Cambridge University Press, ISBN: 0521540518"
}

@article{Belhumeur1999,
  title = {The Bas-Relief Ambiguity},
  volume = {35},
  ISSN = {1573-1405},
  XXurl = {http://dx.doi.org/10.1023/A:1008154927611},
  DOI = {10.1023/a:1008154927611},
  number = {1},
  journal = ijcv,
  publisher = {Springer Science and Business Media LLC},
  author = {Belhumeur,  Peter N. and Kriegman,  David J. and Yuille,  Alan L.},
  year = {1999},
  month = nov,
  pages = {33–44}
}
\end{document}